\pdfoutput=1

\documentclass[11pt]{article}

\usepackage{acl}

\usepackage{times}
\usepackage{latexsym}

\usepackage[T1]{fontenc}

\usepackage[utf8]{inputenc}

\usepackage{microtype}

\usepackage{inconsolata}

\usepackage{graphicx}
\usepackage{color}
\usepackage{amsmath}
\usepackage{booktabs}
\usepackage{multirow}
\usepackage{amssymb}
\usepackage{amsmath}
\usepackage{bm}
\usepackage{makecell}

\title{Two-stage Generative Question Answering on Temporal Knowledge Graph Using Large Language Models}

\newcommand{\blfootnote}[1]{
    \begingroup
    \renewcommand\thefootnote{}\footnote{#1}
    \addtocounter{footnote}{-1}
    \endgroup
}
        
\author{Yifu Gao$^1$, Linbo Qiao$^1$\textsuperscript{*}, Zhigang Kan$^1$, Zhihua Wen$^1$, Yongquan He$^3$, Dongsheng Li$^{1,2}$\textsuperscript{*}\\
        $^1$ National Key Laboratory of Parallel and Distributed Computing, \\
        National University of Defense Technology, Changsha, China\\ 
        $^2$ Xiangjiang Laboratory, Changsha, China\\ 
        $^3$ Meituan, Beijing, China\\
        \texttt{\{gaoyifu, qiao.linbo, kanzhigang13, zhwen,  dsli\}@nudt.edu.cn} \\
        \texttt{heyongquan@meituan.com} }

\begin{document}
\maketitle
\blfootnote{\textsuperscript{*}Corresponding Author}

\begin{abstract}

Temporal knowledge graph question answering (TKGQA) poses a significant challenge task, due to the temporal constraints hidden in questions and the answers sought from dynamic structured knowledge.
Although large language models (LLMs) have made considerable progress in their reasoning ability over structured data, their application to the TKGQA task is a relatively unexplored area.
This paper first proposes a novel \textbf{gen}erative \textbf{t}emporal \textbf{k}nowledge \textbf{g}raph \textbf{q}uestion \textbf{a}nswering framework, GenTKGQA, which guides LLMs to answer temporal questions through two phases: Subgraph Retrieval and Answer Generation. 
First, we exploit LLM's intrinsic knowledge to mine temporal constraints and structural links in the questions without extra training, thus narrowing down the subgraph search space in both temporal and structural dimensions.
Next, we design virtual knowledge indicators to fuse the graph neural network signals of the subgraph and the text representations of the LLM in a non-shallow way, which helps the open-source LLM deeply understand the temporal order and structural dependencies among the retrieved facts through instruction tuning.  
Experimental results on two widely used datasets demonstrate the superiority of our model.

\end{abstract}

\begin{figure}[t]
    \centering
    \includegraphics[width=0.95\columnwidth]{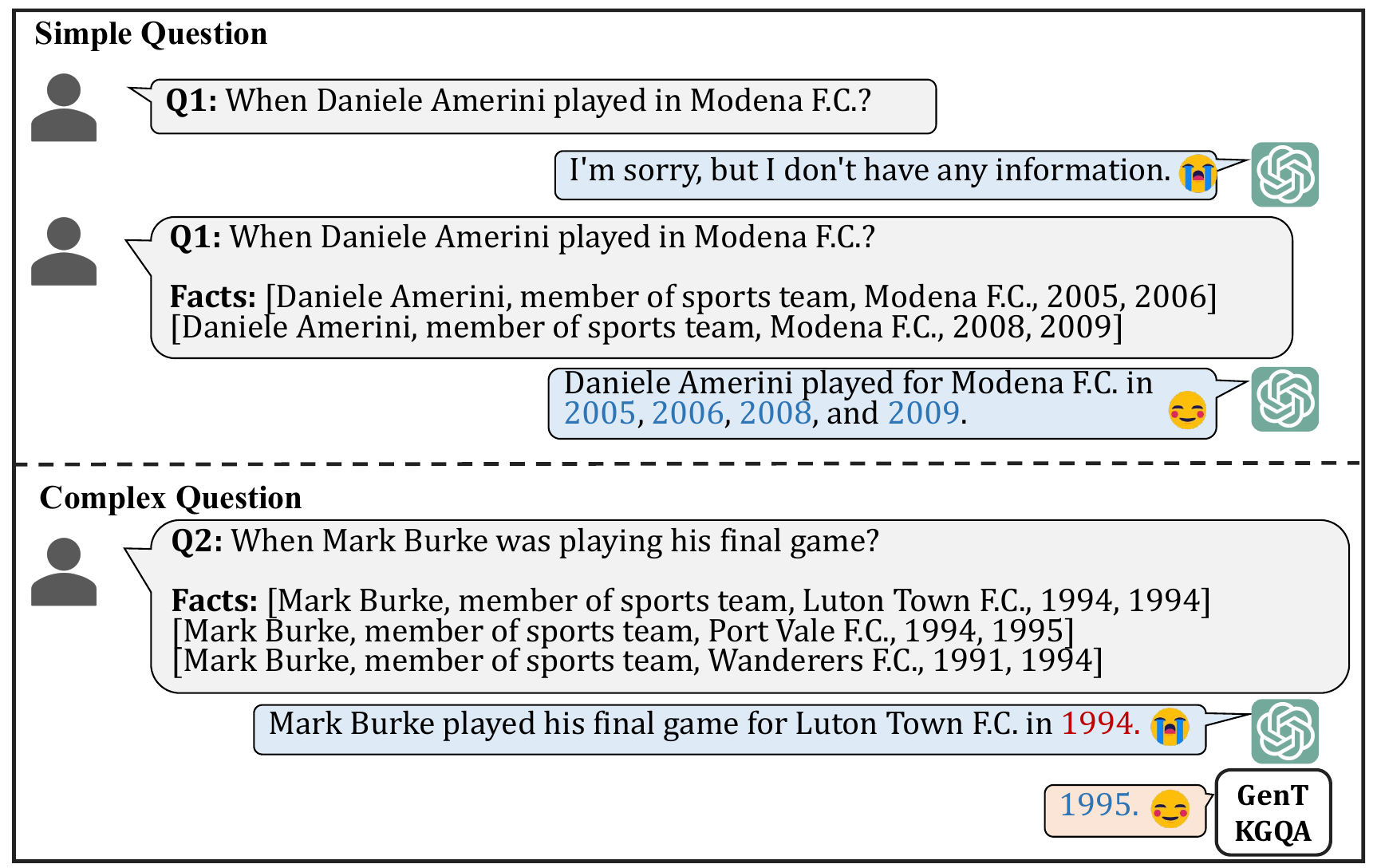}
    \caption{Examples of the responses of LLM and GenTKGQA to the simple and complex temporal questions.}
    \label{fig:intro}
\end{figure}

\section{Introduction}

Real-world knowledge is frequently updated rather than static \cite{wiki,ICEWS}, e.g., (\textit{Obama}, \textit{hold\_position}, \textit{President}) is merely valid only for a certain period [\textit{2009}, \textit{2016}]. Hence, the temporal knowledge graph (TKG) is proposed as a database for storing dynamic structured facts associated with timestamps, denoted as (\textit{subject}, \textit{relation}, \textit{object}, \textit{timestamp}).
Temporal knowledge graph question answering (TKGQA) aims to answer a natural question with explicit or implicit temporal constraints based on the TKG, e.g., "\textit{Who held the position of president (in 2017) or (after Obama)?}". 
Due to the temporal constraints hidden in questions and the answers sought from dynamic structured knowledge, TKGQA is one of the most challenging QA tasks.


Recently, large language models (LLMs) have shown strong competitiveness in various fields \cite{NLP1,NLP2}. Some researchers explore the reasoning capability of LLMs for structured knowledge based on KGQA tasks \cite{zeroshotQA,KG-GPT}, and some works examine the temporal reasoning capabilities of LLMs through time-sensitive QA tasks \cite{timeqa,TEMPReason}. Intuitively, LLMs have the ability to deal with temporal structured knowledge.
Based on the above findings, we attempt to utilize LLMs for the TKGQA task and summarize the following two challenges:
1) \textbf{\textit{Question-relevant Subgraph Retrieval.}} 
A common practice to enhance the LLM's domain-specific reasoning capability is to input query-relevant information as additional knowledge into the LLMs \cite{chatdb}. 
As shown in Figure \ref{fig:intro}, ChatGPT \footnote{https://openai.com/blog/chatgpt} cannot answer temporal questions directly, but it can answer simple questions when the relevant facts are provided.
However, finding facts relevant to the problem is a struggle due to the large search space with both structural and temporal dimensions.
For example, for the complex temporal question "\textit{Who held the position of president after Obama?}", the structural link \textit{hold position} between the entities and the temporal constraints \textit{after 2016} in the problem are unknown, so directly finding all relevant facts about the given entities \textit{Obama} and \textit{President} is bound to introduce too much noise information.
How to accurately retrieve relevant facts from a two-dimensional space is the first challenge.
2) \textbf{\textit{Complex-type Question Reasoning.}}
Recent works about LLM-based KG reasoning mostly input structured knowledge in the natural text form into the task prompt and reason about the answers in a training-free manner \cite{LLMstudy}.
However, these approaches fuse subgraph information with the LLM in a shallow way, which limits the inference performance on the complex question type.
As illustrated in Figure \ref{fig:intro}, ChatGPT cannot understand the chronological order of the relevant facts and answers incorrectly on the complex question type "final".
How subgraph information can be integrated into LLM representations in a non-superficial way to simulate structured reasoning remains an open question.

Hence, we propose GenTKGQA, a novel generative temporal knowledge graph question answering framework consisting of two phases, subgraph retrieval and answer generation, which is used to address the above two challenges, respectively.
At the first phase, we find that the structural and temporal scope of the subgraph is determined by the relation links and the temporal constraints in the question, respectively.
Therefore, we use a divide-and-conquer strategy to reduce the subgraph search space by decomposing the complex subgraph retrieval problem into two subtasks, namely, relation ranking and time mining.
Then, we utilize the LLM's internal knowledge to mine structural connections between entities and time constraints in the problem without extra training.
We only need to input few-shot examples into the prompt to accomplish subgraph retrieval of the entire data.
At the second phase, we fine-tune the open-source LLM with instruction tuning to incorporate structural and temporal information of the subgraph in a non-shallow way.
Recent works illustrate that fusing graph neural network (GNN) representations and language text representations can enhance the ability of LMs to perceive graph structure \cite{GREASELM}.
Thus, we design three novel virtual knowledge indicators to bridge the links between pre-trained GNN signals of the temporal subgraph and text representations of the LLM, which guides the LLMs in deeply understanding the graph structure and improves their reasoning ability for complex temporal questions.
Overall, our contribution can be summarized in the following four points:

1) We present a novel two-stage generative framework for the TKGQA task, which explores LLM's temporal reasoning capabilities in the context of dynamic structured knowledge.

2) We motivate the LLM's intrinsic knowledge to mine the temporal constraints and structural connections in the questions without extra training, which reduces the subgraph search space from both structural and temporal dimensions. 

3) We design virtual knowledge indicators to fuse the GNN signals and text representations in a non-shallow way, which helps the open-source LLMs improve their reasoning on the complex question type through instruction tuning.  

4) Experiment results on two widely used datasets show that GenTKGQA as a generative model performs consistently better than embedding-based methods on the Hits@1 metric.

\section{Related Work}


\subsection{TKGQA Methods}

Temporal knowledge graph question answering (TKGQA) task aims to answer complex questions in the natural language format using entities and timestamps from the given TKG \cite{TempQuestions,TimeQuestions,multiqa}.
Existing mainstream methods employ TKG embeddings to represent the entities, relations and timestamps, and use the scoring function to select the entity or time with the highest relevance as the answer \cite{cronqa}.
However, single embedding methods have difficulty handling complex reasoning problems with implicit time constraints. Therefore, recent methods try to incorporate other modules to improve the model performance on complex problems.
Specifically, TSQA \cite{TSQA} proposes a contrastive approach to enhance the model's time sensitivity.
TempoQR \cite{tempoqa} designs three modules, namely context, entity and time-aware information, to enhance the incorporation of the TKG into questions.
Besides, some approaches propose to solve the TKGQA task with problem-relevant subgraph reasoning \cite{subgraphreasoning,LGQA,TwiRGCN}.
Despite the effectiveness of these approaches, few studies have explored how LLMs can address the TKGQA task.



\subsection{LMs for Temporal Question Answering}

Language models (LMs) have exhibited strong performance on the question answering task \cite{QAbenchmark2}.
In recent years, some researchers have explored the temporal reasoning capabilities of LMs and propose several typical time-sensitive QA datasets.
They focus on temporal question answering either within a closed-book setting to assess models' internal memorization of temporal facts \cite{streamingqa,templama}, or within an open-book setting to evaluate models' temporal understanding and reasoning capability over unstructured texts \cite{situatedqa,timeqa,TEMPReason}.
In the context of the latter setting, some works propose to use the graph structure extracted from text to assist the model in determining the temporal order between events \cite{DocTime,fusetransformer,rememo,TGQA}, which is similar but fundamentally different from our work. 
These approaches aim to answer temporal questions based on the known natural text context. In contrast, our model focuses on structured temporal knowledge as auxiliary information that needs to be retrieved by the model.

\subsection{LMs for KG Question Answering}

How to combine LMs and KG for question answering has become a hot issue. Some works attempt to enhance question representation and relation matching with PLMs in the multi-hop KGQA task \cite{EmbeddedKGQA,subgraphenhanced,unikgqa}, but there is no interaction between the LM and KG representations. Other works try to use one modality to ground the other, i.e., using the encoded representation of a linked KG to augment the text representation \cite{KagNet,KTNET}, or using the text representation of the PLM to enhance the graph reasoning model \cite{Knowledge-AwareQA}.
The most recent approaches enable deep integration of the two modalities by jointly updating the GNN and LM representations \cite{QAGNN, GREASELM}.

However, the emergence of large language models (LLMs) has changed how LMs handle the KGQA task, which is divided into two main approaches: training-free and fine-tuning \cite{LLMstudy}.
Recent works attempt to append query-relevant facts as the input prompt for LLMs and make inferences without extra training \cite{zeroshotQA,RRA,structgpt,KG-GPT,COK}. 
Fine-tuning the full parameters of the LLM can be cost-prohibitive.
Hence, KPE \cite{KPE} enables knowledge integration by freezing PLM parameters and introducing trainable parameter adapters.
ChatKBQA \cite{ChatKBQA} employs the LoRA \cite{lora} technique to fine-tune open-source LLMs, achieving the logical query form generation.
Besides, KoPA  \cite{KoPA} incorporates the KG embeddings into the LLMs with a prefix adapter, aiming to achieve structrual-aware reasoning in the LLMs. 
Applying LLMs to the temporal KGQA task remains an unexplored area.

\section{Preliminaries}
\paragraph{TKGQA.}
A temporal knowledge graph (TKG) $\mathcal{G} := (\mathcal{E}, \mathcal{R},\mathcal{T}, \mathcal{F})$ is a multi-relational, directed graph with timestamped edges between entities, where $\mathcal{E}$, $\mathcal{R}$ and $\mathcal{T}$ represent the sets of entities, relations and timestamps, respectively. 
Each fact in the $\mathcal{G}$ can be represented as a quadruple $(s,r,o,t) \in \mathcal{F} $, corresponding to entity $s/o \in \mathcal{E} $, relation type $r \in \mathcal{R}$ and timestamp $t \in \mathcal{T} $. 
Given a natural language question $q$, TKGQA aims to extract entities $s/o$ or timestamps $t$ that correctly answer the question $q$.
\paragraph{ICL and IT.}
Applying LLM to the TKGQA task, the goal is to generate the answer $\mathcal{A}$ based on the input text sequence $\mathcal{S}$ and the LLM $\mathcal{M}$.
$\mathcal{S}$ consists of several parts: the instruction prompt $\mathcal{I}$, the task-specific input prompt $\mathcal{Q}$, and the auxiliary demonstration prompt $\mathcal{D}$.
In-context Learning (ICL) method is an efficient approach to employ LLMs to solve downstream tasks without extra training, the input sequence of ICL can be denoted as $\mathcal{S} = \mathcal{I}:\mathcal{D}:\mathcal{Q}$, where $:$ means to concatenate the different prompts.
Meanwhile instruction tuning (IT) aims to fine-tune LLMs to follow human instructions and accomplish the distinct tasks in the instruction prompt, the input sequence of IT can be denoted as $\mathcal{S} = \mathcal{I}:\mathcal{Q}:\mathcal{A}$.

\section{Method}

We apply the LLMs processing TKGQA task in a two-phase process, i.e., an ICL-based subgraph retrieval phase and an IT-based answer generation phase.
At the first stage, we utilize internal knowledge of LLM for unlabeled subgraph retrieval. 
At the second stage, we incorporate external knowledge for structure-aware temporal inference.

\begin{figure*}
    \centering
    \includegraphics[width=0.99\textwidth]{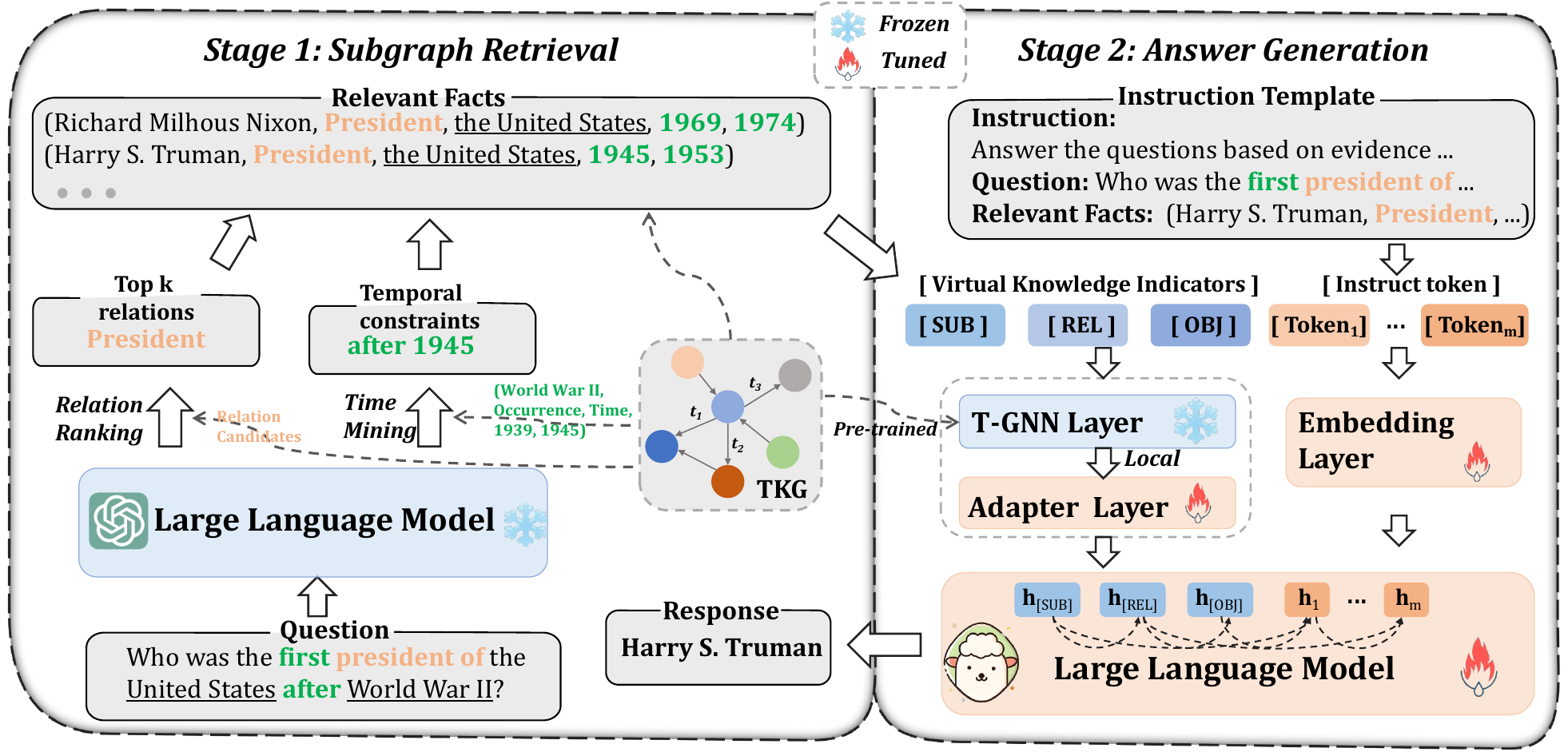}
    \caption{The overall architecture of our proposed GenTKGQA can be divided into two stages, subgraph retrieval and answer generation. Given a temporal question, we mark the entities provided, the implied time constraints and the links between the entities with \underline{underline}, \textcolor{blue}{blue font} and \textcolor{orange}{orange font}, respectively.}
    \label{fig:model}
\end{figure*}

\subsection{Subgraph Retrieval} \label{sec:SR}

We split the complex subgraph retrieval problem into a relation ranking subtask in the structural dimension and a time mining subtask in the temporal dimension.
With such a strategy, we only need to provide a small number of samples to complete the subgraph retrieval.

\subsubsection{Relation Ranking}\label{sec:rr}

We aim to determine the structural scope of the subgraph $\mathcal{G}_{sub,i}$ by retrieving the corresponding relations from the candidate  relation set $\mathcal{R}_i$ for each question $q_i$.
Recent work has shown that LLMs can better handle the information extraction task as re-ranking agents \cite{rankgpt}.
Therefore, we feed the question $q_i$ and the candidate set $\mathcal{R}_i$ to the LLM to obtain the top $k$ relations $\mathcal{R}_{i,k}$ relevant to the question.
Relations in $\mathcal{R}_i$ are linearized, i.e., [employer, member\_of\_sports\_team, ..., ]), and the retrievd relations can bridge the entities identified within the questions.
The specific relation ranking prompt is shown in Appendix \ref{appendix:prompt}.

\subsubsection{Time Mining}

We find that natural questions contain temporal constraints, either explicit or implicit, and we can easily determine the range of relevant facts in the temporal dimension by using explicit temporal constraints such as "in 2008", "at the year of 2012", etc.
How to capture implicit temporal constraints is the key to improving the efficiency of searching for relevant facts.
For example, for the complex temporal question "\textit{Who held the position of president after Obama?}", the implicit temporal restriction is known to be (\textit{after 2016}) based on the temporal validity (\textit{2009}, \textit{2016}) of the fact (\textit{Obama}, \textit{hold\_position}, \textit{President}). 
We design specific prompt templates based on different answer types as well as consider the temporal validity of the facts in the questions (given entities and relations matched in the Section \ref{sec:rr}) to get the temporal constraints. The details of the templates are shown in Appendix \ref{appendix:prompt}.

Through the above process, we narrow down the search space of subgraphs and use relevant facts under structural and temporal constraints as additional knowledge to assist the LLM inference.

\subsection{Answer Generation}

In this section, we will discuss how to incorporate the knowledge retrieved in the previous section \ref{sec:SR} into the LLM.
Previous fundamental approaches to incorporate KG structural information focus on adding the knowledge to the input prompt in the text form, i.e., (subject, relation, object).
However, incorporating query-relevant facts into LLMs in text form is not a good choice. 
Because such shallow interactions do not enable the model to understand the structural dependencies and temporal order between facts, leading to weak temporal reasoning in the complex problem type.

Recent works show that language models can enhance their ability to perceive graph structures by incorporating knowledge representations expressed by graph neural networks (GNN) \cite{GREASELM}.
Inspired by this, we first extract the structural and temporal information of entities and relations with pre-trained temporal GNN embeddings.
Then, we bridge the links between GNN and text representations through pre-designed virtual knowledge indicators.
At last, we fine-tune the open-source LLM to deeply understand the temporal order and structural dependencies of the retrieved query-relevant facts.


\subsubsection{Temporal GNN}

Given the retrieved temporal subgraph $\mathcal{G}_{sub,z}$ of question $q_z$, we first initialise entity, relation and time representations in $\mathcal{G}_{sub,z}$ using the TKG embedding method \cite{TNTComplEx}.
Then, to fully explore the structural information among entities and relations of the temporal subgraph, we propose a temporal graph neural network (T-GNN), which is a variant of graph attention networks \cite{gat}.
The important distinction between them is that T-GNN captures the correlation scores of neighbouring nodes by incorporating temporal embeddings.
Therefore, T-GNN computes message $\mathbf{m}_{ij}$ between entities $e_i$ and $e_j$ as follows:
\begin{equation}
\mathbf{m}_{ij}=\mathbf{W}_{m}(\mathbf{e}_{i}^{(l-1)}+\mathbf{r}_{ij}+\mathbf{t}_{ij}),
\end{equation}
where $\mathbf{e}_i^{(l-1)}$ is the entity representation of $e_i$ at the $l$-1 layer, $\mathbf{r}_{ij}$ and $\mathbf{t}_{ij}$ are the embeddings of the relation and timestamp connecting $e_i$ and $e_j$. $\mathbf{W}_{m}$ is a linear transformation.
Next, the node representation $\mathbf{e}_{j}^{(l)}$ is calculated via message passing between neighbors on the $\mathcal{G}_{sub,z}$:
\begin{equation}\label{eq:2}
\mathbf{e}_{j}^{(l)}=\sum_{i \in \mathcal{N}_{j}} \alpha_{i j} \mathbf{m}_{i j},
\end{equation}
here $\mathcal{N}_{j}$ represents the neighbor entities of the arbitrary node $e_{j}$, and $\alpha_{i j}$ denotes the attention values with $\mathbf{e}_{i}^{(l-1)}$ as query and $\mathbf{m}_{i j}$ as key:
\begin{equation}
\alpha_{i j}=\frac{\exp \left(u_{i j}\right)}{\sum_{w \in \mathcal{N}_{j}} \exp \left(u_{w j}\right)},
\end{equation}
\begin{equation}
u_{i j}=f_n((\mathbf{W}_{q} \mathbf{e}_{i}^{(l-1)})^{\top}\left(\mathbf{W}_{k} \mathbf{m}_{i j}\right)),
\end{equation}
$\mathbf{W}_{q}, \mathbf{W}_{k}$ are linear transformations, $f_n$ is the RELU activation function.
Through the above process, we obtain the entity representation $\mathbf{e}_j$ with subgraph structural and temporal information.
Following embedding-based TKGQA methods, we use the link prediction task to pre-train the graph neural network representations. Specifically, for each fact $(s,r,o,t)$ in the TKG, we generate a query $(s,r,[mask],t)$ or $([mask],r,o,t)$ by masking  the  object or subject entity.  Then, we obtain the mask embedding $\mathbf{e}_{[mask]}^{(l)}$ through Eq.(\ref{eq:2}),  and feed it into the multi-layer perceptron (MLP) decoder to maximize the probability of the missing entity $o$  and $s$ through the cross-entropy loss function:
\begin{equation}
    p\left(\mathbf{e} \right) = \text{Softmax} (\mathbf{e}_{[mask]}^{(l)}\mathbf{w}+\mathbf{b}),
\end{equation}
\begin{equation}
  \mathcal{L}=-\sum_{(s, r, o, t)\in\mathcal{G}} \log p({o_t})+\log p({s_t}).
\end{equation}

\subsubsection{Virtual Knowledge Indicators}

We design three knowledge indicators to link graph signals and input prompt text, namely [SUB], [REL] and [OBJ], correspond to the virtual tokens of the head entities, relations and tail entities in the subgraph, respectively. 
We then try to incorporate structural and temporal information from the subgraph into the indicator representations.
Specifically, we use the $\operatorname{Local}$ operator to get the structure representations $\mathbf{h}^{s}$ of entities and relationships in the subgraph, respectively:
\begin{equation}
\mathbf{h}_{\operatorname{[SUB]}}^{s}=\operatorname{Local}(\mathbf{e}_{\operatorname{[SUB]}}),
\end{equation}
here $\mathbf{e}_{\operatorname{[SUB]}}$ represents pre-trained T-GNN embeddings of all subject entities in the subgraph, $\operatorname{Local}$ indicates the  max or mean pooling operator.
Besides, we leverage the time embeddings to enhance the indicator representations with temporal information. 
\begin{equation}\label{eq:6}
\mathbf{h}_{\operatorname{[SUB]}}^{st}=\mathbf{h}_{\operatorname{[SUB]}}^{s}+\mathbf{t}_{min}+\mathbf{t}_{max},
\end{equation}
where $\mathbf{t}_{min}$ and $\mathbf{t}_{max}$ denote the embeddings for the minimum and maximum values of time in the subgraph, respectively. 
The intuition follows BERT that use position embeddings for tokens \cite{BERT}. Here, time embeddings can be seen as entity positions in the time dimension.
The relation and object indicators are the same as subject.
At last, we employ a simple linear layer $\mathbf{W}_{p}$ to project them into the textual representation space of the LLM. The final input prompt sequence $\mathcal{S} = \mathcal{V}:\mathcal{I}:\mathcal{Q}:\mathcal{A}$, $\mathcal{V}$ represent virtual indicator tokens.
Details of the instruction template can be found in the Appendix \ref{appendix:prompt}.
The optimization objective of the LLM $\mathcal{M}$ can be formulated as:
\begin{equation}
\mathcal{A}=\arg \max _{\mathcal{A}} P_{\mathcal{M}}\left(\mathcal{A} \mid \mathcal{V}, \mathcal{I}, \mathcal{Q}, \mathcal{A}\right).
\end{equation}

\begin{table*}[ht]
\centering
    \resizebox{\textwidth}{!}{
    \begin{tabular}{l|c|c|c|c|c|c|c|c|c|c}  
    \toprule
    \multirow{3}{*}{\textbf{Model}} & \multicolumn{5}{c|}{\textbf{Hits@1}}&\multicolumn{5}{c}{\textbf{Hits@10}} \\ 
    \cline{2-11}
    & \multirow{2}{*}{\textbf{Overall}} & \multicolumn{2}{c|}{\textbf{Question Type}} & \multicolumn{2}{c|}{\textbf{Answer Type}} & \multirow{2}{*}{\textbf{Overall}} &\multicolumn{2}{c|}{\textbf{Question Type}} & \multicolumn{2}{c}{\textbf{Answer Type}} \\  
    \cline{3-6} 
    \cline{8-11}
    & & \textbf{Complex} & \textbf{Simple} & \textbf{Entity} & \textbf{Time} &  &\textbf{Complex} & \textbf{Simple} & \textbf{Entity} & \textbf{Time} \\
    \cline{1-11}
    EmbedKGQA & 0.288 & 0.286 & 0.290 & 0.411 & 0.057 & 0.672 & 0.632 & 0.725 & 0.850 & 0.341\\
    EaE & 0.288 & 0.257 & 0.329 & 0.318 & 0.231 & 0.678 & 0.623 & 0.753 & 0.668 & 0.698\\
    \cline{1-11}
    CronKGQA & 0.647 & 0.392 & 0.987 & 0.699 & 0.549 & 0.884 & 0.802 & 0.990 & 0.898 & 0.857\\
    EntityQR & 0.745 & 0.562 & 0.990 & 0.831 & 0.585 & 0.944 & 0.906 & 0.993 & 0.962 & 0.910\\
    TMA & 0.784 & 0.632 & 0.987 & 0.792 & 0.743 & 0.943 & 0.904 & 0.995 & 0.947 & 0.936\\
    TSQA & 0.831 & 0.713 & 0.987 & 0.829 & 0.836 & \underline{0.980} & \underline{0.968} & \underline{0.997} & \textbf{0.981} & \underline{0.978}\\
    TempoQR & \underline{0.918} & \underline{0.864} & \underline{0.990} & \underline{0.926} & \underline{0.903} & 0.978 & 0.967 & 0.993 & \underline{0.980} & 0.974\\
    \cline{1-11}
    BERT \textit{w/o tkg} & 0.071 & 0.086 & 0.052 & 0.077 & 0.06 & 0.213 & 0.205 & 0.225 & 0.192 & 0.253\\
    RoBERTa \textit{w/o tkg}& 0.07 & 0.086 & 0.05 & 0.082 & 0.048 & 0.202 & 0.192 & 0.215 & 0.186 & 0.231\\
    ChatGPT \textit{w/o tkg} & 0.151  & 0.144 & 0.160 & 0.134 &0.182  & 0.308 & 0.308 & 0.307 & 0.257 & 0.402\\
    \cline{1-11}
    BERT \textit{w/ tkg} & 0.243 & 0.239 & 0.249 & 0.277 & 0.179 & 0.620 & 0.598 & 0.649 & 0.628 & 0.604\\
    RoBERTa \textit{w/ tkg} & 0.225 & 0.217 & 0.237 & 0.251 & 0.177 & 0.585 & 0.542 & 0.644 & 0.583 & 0.591\\
    ChatGPT \textit{w/ tkg} & 0.754  & 0.579 & 0.987 & 0.689 &0.873&0.852  & 0.746 & 0.992 & 0.808 & 0.933\\
    \cline{1-11}
    GenTKGQA & \textbf{0.978}&\textbf{0.962}&\textbf{0.999}&\textbf{0.967}&\textbf{0.990}&\textbf{0.983}&\textbf{0.971}&\textbf{0.999}&0.974&\textbf{0.994}\\
    \bottomrule
    \end{tabular}}
    \caption{Performance comparison of different models on CronQuestions. The best and second best results are marked in \textbf{bold} and \underline{underlined}, respectively.
    \textit{w/o tkg} indicates that LMs answer the questions directly without using TKG information, and \textit{w/ tkg} indicates that LMs answer the questions with TKG background knowledge.}
\label{tab:main result}
\end{table*}

\section{Experiments}

We design experiments to answer the following questions: 

\textbf{Q1.}How does GenTKGQA perform on the TKG question answering task? (Section \ref{sec:mainresult})

\textbf{Q2.}How do the two stages contribute to the model performance respectively? (Section \ref{sec:ablation})

\textbf{Q3.}How does GenTKGQA perform under changes in hyper-parameters? (Section \ref{sec:sensity})

\textbf{Q4.}How does GenTKGQA outperform ChatGPT in answering complex temporal questions? (Section \ref{sec:qualitative})

\subsection{Datasets, Metrics and Baselines}

CronQuestions \cite{cronqa} is a temporal QA dataset, which contains 410K unique question-answer pairs, including annotated entities and timestamps, with 350k for training and 30k for validation and testing. The dataset can be categorized into simple reasoning (Simple Entity and Simple Time) and complex reasoning (Before/After, First/Last, and Time Join) based on temporal constraints. 
The TimeQuestions dataset \cite{TwiRGCN}  has 13.5k manually edited questions and is divided into three parts: training, validation, and testing, containing 7k, 3.2k, and 3.2k questions, respectively.
The questions are categorized into four types: Explicit, Implicit, Temporal and Ordinal.
Following previous studies, we use two popular evaluation metrics, Hits@1 and Hits@10.
More information about datasets and metrics can be found in Appendix \ref{appendix:dataset}.

We compare four types of baselines: 1) KG embedding-based models including EaE \cite{EaE} and EmbedKGQA \cite{EmbeddedKGQA}; 2) TKG embedding-based models including CronKGQA \cite{cronqa}, EntityQR \cite{tempoqa}, TMA \cite{TMA}, TSQA \cite{TSQA} and TempoQR \cite{tempoqa}; and 3) Language models, including BERT \cite{BERT}, RoBERTa \cite{roberta}, and ChatGPT. 
For another TimeQuestions dataset, we use CronKGQA, TempoQR and TwiRGCN \cite{TwiRGCN} for comparison.
The implementation details of the baselines are described in Appendix \ref{appendix:baseline}.

\begin{table}
\centering
\resizebox{\linewidth}{!}{
    \begin{tabular}{l|ccccc}  
    \toprule
    \textbf{Model} & \textbf{Overall} & \textbf{Explicit} & \textbf{Implicit} & \textbf{Temporal} & \textbf{Ordinal}\\
    \midrule
    CronKGQA & 0.462 & 0.466 & 0.445 & 0.511 & 0.369 \\
    TempoQR	& 0.416	& 0.465	& 0.360 & 0.400 & 0.349 \\
    TwiRGCN(average) & \textbf{0.605} & \textbf{0.602} & 0.586 & 0.641 & 0.518\\
    TwiRGCN(interval) & 0.603 & 0.599 & 0.603 & \textbf{0.646} & 0.494\\
    GenTKGQA & 0.584 & 0.596 & \textbf{0.611} & 0.563& \textbf{0.578}\\
    \bottomrule
    \end{tabular}}
    \caption{Hits@1 for different models on TimeQuestions.}
\label{tab:mainresult2}
\end{table}

\subsection{Main Results} \label{sec:mainresult}

Table \ref{tab:main result} reports the performance of all methods on the CronQuestions dataset for various question types. 
We can observe that GenTKGQA consistently outperforms the baselines in terms of "Overall" performance, and achieves significant improvements of 11.3\% in the "Complex" question type and 9.6\% in the "Time" answer type on the Hits@1 metric over the second best method.
Especially, our model achieves nearly 100\% for the "Simple" question type.
The possible reason is that simple questions usually involve single facts, GenTKGQA can easily retrieve the relevant facts containing the answer and infer the correct answer through instruction tuning technique.

Furthermore, compared to KG embedding methods, temporal KG embedding methods show significant results on various metrics, thanks to the fact that the temporal information of the TKG is taken into account in the question representation. 
This is also why KG embedding methods are particularly ineffective for the "Time" answer type.
However, most TKG embedding methods treat the QA task as a link prediction, which works for the "Simple" question type containing a single fact compared to the "Complex" question type.

We find that PLMs (BERT, RoBERTa) and LLMs (ChatGPT) have the lowest performance on the TKGQA task without TKG information. This suggests that language models (LM), whether encoded or generated, with a large or small number of parameters, have difficulty answering temporal questions without any relevant context. 
\textit{w/ tkg} indicates that LMs use entity/time embeddings or relevant facts from the TKG. 
Obviously, LMs \textit{w/ tkg} have significantly better performance, which suggests that the language models have some degree of temporal reasoning capability when relevant TKG information is provided, validating the importance of the subgraph retrieval phase. 
It is worth noting that ChatGPT \textit{w/ tkg} still performs weakly in reasoning about complex problem types when providing the facts retrieved in the first stage by GenTKGQA, while our model achieves the best results.
This demonstrates the effectiveness of interacting GNN and LM representations in dealing with complex temporal problems.
The above findings demonstrate the adequacy of our two motivations for solving complex temporal problems with LLMs.
A possible reason for the poor improvement of our model's Hits@10 metric for the "Entity" type is that the LMs provide irrelevant responses when asked to generate multiple answers.

As shown in Table \ref{tab:mainresult2}, GenTKGQA still achieves surprising results on the TimeQuestions dataset, especially improving on the "Implicit" and "Ordinal" question types by 1.3\% and 11.6\%, respectively.
Both question types have implied time constraints, similar to the "Complex" question type on the CronQuestions dataset.
Overall, compared to the baseline methods, GenTKGQA achieves significant results on the complex question type with different datasets, validating the motivation of our work and the effectiveness of the proposed model.
Last but not least, GenTKGQA, as a generative QA model, achieves better results than most traditional extractive QA methods.

\begin{table}
\centering
	\resizebox{.95\linewidth}{!}{
    \begin{tabular}{l|c|c|c|c|c}  
    \toprule
    \multirow{3}{*}{\textbf{Model}} & \multicolumn{5}{c}{\textbf{Hits@1}} \\ 
    \cline{2-6}
    & \multirow{2}{*}{\textbf{Overall}} & \multicolumn{2}{c|}{\textbf{Question Type}} & \multicolumn{2}{c}{\textbf{Answer Type}}  \\  
    \cline{3-6} 
    & & \textbf{Complex} & \textbf{Simple} & \textbf{Entity} & \textbf{Time} \\
    \cline{1-6}
    GenTKGQA &  0.978 & 0.962 & 0.999 & 0.967 & 0.990  \\
    \textit{w/o SR}  & 0.119 & 0.140 & 0.090 & 0.127 & 0.103 \\
    \textit{w/o SR inference} &  0.475 & 0.381 & 0.601 & 0.294 & 0.812\\
    \textit{w/ SR random} &  0.766 & 0.613 & 0.970 & 0.661 & 0.961\\
    \textit{w/o T-GNN} &  0.935 & 0.914 & 0.965 & 0.920 & 0.963\\
    \textit{w/o VKI} &  0.843 & 0.824 & 0.870 & 0.831 & 0.867\\
    \bottomrule
    \end{tabular}}
    \caption{Ablation study results on CronQuestions.}
\label{tab:ablation study}
\end{table}

\begin{table*}[t]
\centering
\resizebox{\linewidth}{!}{
    \begin{tabular}{c|c|c|c}  
    \toprule
    \multirow{2}{*}{\textbf{Type}}&\textbf{Question/}&\multicolumn{2}{c}{\textbf{Response}}\\
    \cline{3-4}
    & \textbf{Retrieved Graph}&\textbf{ChatGPT \textit{w/ tkg}} &\textbf{GenTKGQA}\\
    \cline{1-4}
    \multirow{6}{*}{\textbf{Simple}} &\textbf{dean in 1997 was the person?}& &\\
    \multirow{6}{*}{\textbf{Entity}} &   [Xavier Darcos, position held, dean, 1995, 1998]&[ \textcolor{blue}{Katarzyna Olbrycht}, \textcolor{blue}{José Miguel Pérez García},&[ \textcolor{blue}{Xavier Darcos},  \textcolor{blue}{Zinaida Belykh},\\
    &[Zinaida Belykh, position held, dean, 1988, 1998]&  \textcolor{blue}{Jiří Zlatuška}, \textcolor{blue}{Xavier Darcos},& \textcolor{blue}{José Miguel Pérez García},  \textcolor{blue}{Jiří Zlatuška},\\
     &[José Miguel Pérez García, position held, dean, 1990, 1998]&  \textcolor{blue}{Zinaida Belykh}, Andrei Fursenko, & \textcolor{blue}{Katarzyna Olbrycht}, Catalina Enseñat Enseñat,\\
     & [Jiří Zlatuška, position held, dean, 1994, 1998]& Anatoly Torkunov, Alexander Konovalov, &Catalina Enseñat Enseñat, Marcel Berger,\\
      &[Katarzyna Olbrycht, position held, dean, 1981, 1998]& Anatoly Vichnyakov, Anatoly Vishnevsky]&Miguel Beltrán Lloris, Miklós Réthelyi]\\
      \midrule
      \multirow{3}{*}{\textbf{Simple}}&\textbf{When Daniele Amerini played in Modena F.C.?}&&\\
    \multirow{3}{*}{\textbf{Time}}&  [Daniele Amerini, member of sports team, Modena F.C., 2005, 2006]&[\textcolor{blue}{2005}, \textcolor{blue}{2006},&[\textcolor{blue}{2005}, \textcolor{blue}{2006},\\
    &[Daniele Amerini, member of sports team, Modena F.C., 2008, 2009]&\textcolor{blue}{2008}, \textcolor{blue}{2009}]&\textcolor{blue}{2008}, \textcolor{blue}{2009}]\\
    

 \midrule
    \multirow{6}{*}{\textbf{Before/}}&\textbf{Who held the position of Sociétaire of the }& &\\
    \multirow{6}{*}{\textbf{After}}& \textbf{ Comédie-Française before Catherine Hiegel?}&&\\
    &[François Jules, position held, Comédie-Française, 1850, 1894]& &\\
     &[Jean Martinelli, position held, Comédie-Française, 1930, 1950]&[Lise Delamare]&[\textcolor{blue}{Yvonne Gaudeau}]\\
     & [Yvonne Gaudeau, position held, Comédie-Française, 1950, 1986]&&\\
     &[Lise Delamare, position held, Comédie-Française, 1951, 1967]&&\\
    
      \midrule
      \multirow{5}{*}{\textbf{First/}}&\textbf{When Mark Burke was playing his final game?}& &\\
    \multirow{5}{*}{\textbf{Last}}&  [Mark Burke, member of sports team, Luton Town F.C., 1994, 1994]&&\\
    &[Mark Burke, member of sports team, Port Vale F.C., 1994, 1995]& [1994]&[\textcolor{blue}{1995}]\\
     &[Mark Burke, member of sports team, Wanderers F.C., 1991, 1994]&&\\
     & [Mark Burke, member of sports team, Darlington F.C., 1990, 1990]&&\\
     
      \midrule
     \multirow{5}{*}{\textbf{Time}}&\textbf{Who was Iowa Governor in Greater German Reich }& &\\
    \multirow{5}{*}{\textbf{Join}}&  \textbf{during the World War II?}&& \\
    &[Bourke B. H., position held, Governor of Iowa, 1943, 1945]&[\textcolor{blue}{George A. W.}]&[\textcolor{blue}{George A. W.},\\
     &[George A. W., position held, Governor of Iowa, 1939, 1943]&&\textcolor{blue}{Bourke B. H.}]\\
     &[Robert D. B., position held, Governor of Iowa, 1945, 1949]& &\\

    \bottomrule
    \end{tabular}}
    \caption{Comparison of responses to five different question types between our GenTKGQA and ChatGPT \textit{w/ tkg}. Marked in blue is the correct answer.}
\label{tab:case study}
\end{table*}

    

\subsection{Ablation Study}\label{sec:ablation}

As shown in Table \ref{tab:ablation study}, to verify each module's importance, we conduct ablation experiments on the CronQuestions dataset.

\textit{w/o SR} means that we directly perform problem inference without using relevant subgraph information in the model training and inference phases, while \textit{w/o SR inference} means that we do not provide subgraphs only at the inference.
We can observe a sharp decrease in model effectiveness due to the lack of use of subgraph information, which is consistent with the other LMs (Section \ref{sec:mainresult}). 
This result shows that current LMs are weak in dealing directly with temporal reasoning problems, validating the importance of the subgraph retrieval module.
The \textit{w/o SR inference} result indicates that GenTKGQA remembers part of the structured knowledge during the training phase and improves the temporal inference performance without providing subgraph information.
\textit{w/ SR random} denotes the random selection of relevant facts involving entities in the question, and the drop in results proves the validity of our first-stage approach.

\textit{w/o T-GNN} indicates that we directly use the temporal embeddings \cite{TNTComplEx} to represent entities and relations of subgraphs. The slight decrease in the results indicates that the T-GNN is able to perceive the structural information of the TKG.
\textit{w/o VKI} means that we try to remove the virtual knowledge indicators from the input prompt. 
Model performance degradation shows that indicators can bridge the gap between distinct representations.

\begin{figure}[t]
    \centering
    \includegraphics[width=0.99\columnwidth]{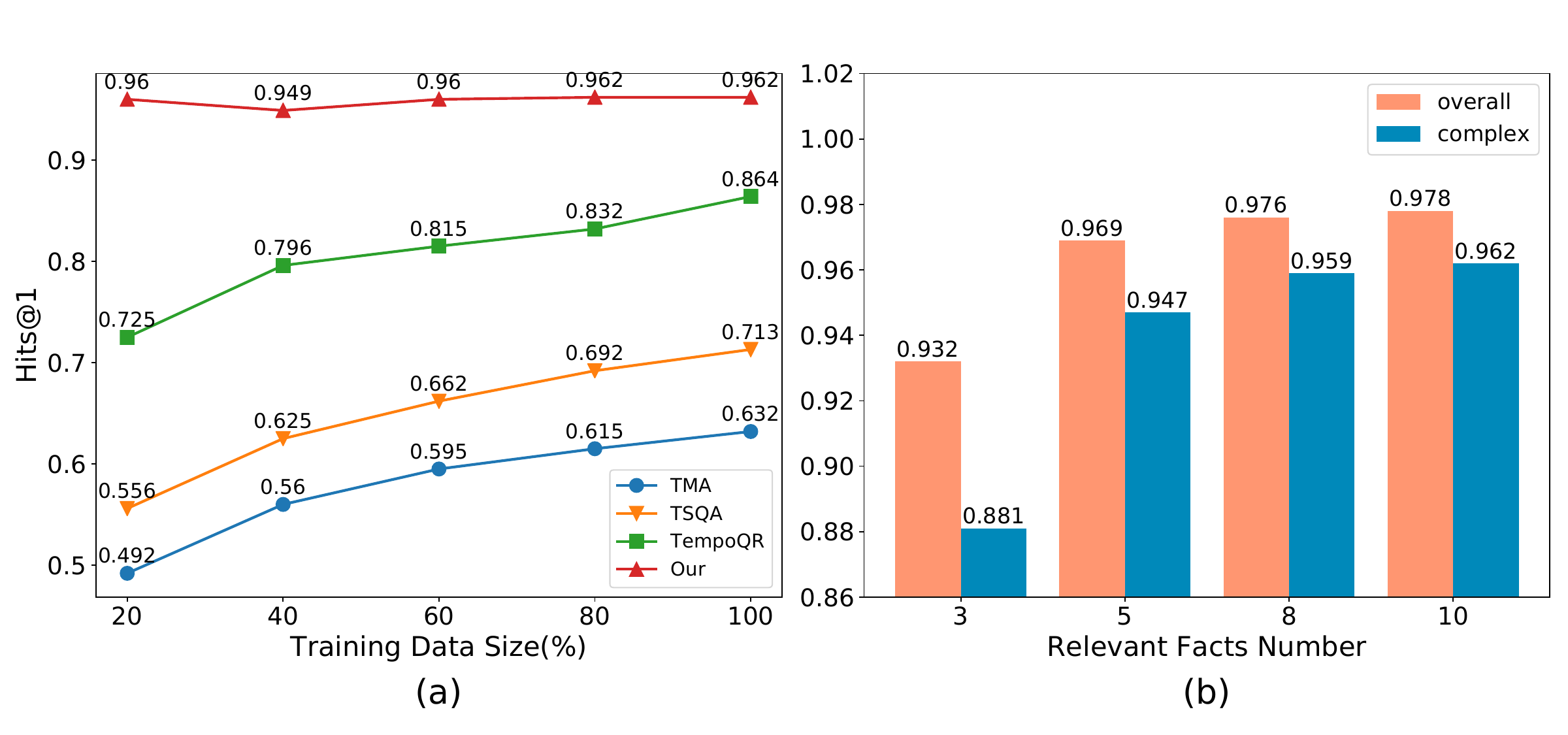}
    \caption{Parameter sensitivity on our GenTKGQA.}
    \label{fig:sensitivity}
\end{figure}

\subsection{Sensitivity Analysis} \label{sec:sensity}

\paragraph{Impact of training data size.}

We explore the impact of different training data sizes to reason about complex temporal questions.
As shown in Figure \ref{fig:sensitivity}(a), by comparing the Hits@1 metric of several methods for the complex question type, our method consistently outperforms others as the training data expands.
In particular, at 20\% of the training data, our model outperforms the second best model by 32\%, demonstrating that our model has strong inference ability in the case of few-shot samples due to its intrinsic knowledge.

\paragraph{Impact of the number of relevant facts.}
We report the performance changes on the CronQuestions dataset by varying the number of retrieved facts $n$ in Figure \ref{fig:sensitivity}(b).
It can be seen that the model performs poorly with a small number of relevant facts ($n$=3), and there is a slight increase in performance at $n$=10. Fewer facts do not provide sufficient context knowledge, while more facts may introduce noise. Taking this into consideration, we set the hyper-parameter $n$ to 10.

\subsection{Qualitative Results} \label{sec:qualitative}

We provide specific examples for each question type to compare the answer results of ChatGPT and ChatGPT \textit{w/ tkg}.
Table \ref{tab:case study} includes the graphs retrieved by our method, along with the answer results for five different question types.

When providing relevant facts retrieved by GenTKGQA as background knowledge, ChatGPT performs competitively in the simple question type, correctly answering questions with entity or time as the answer. However, it has difficulty answering complex types of questions. 
For example, in the "Before/After" and "First/Last" question types, ChatGPT struggles to understand the temporal order of the relevant facts and gives incorrect answers. 
Besides, in the case of the "Time Join" question type, some of the correct answers are missing from the generated responses because ChatGPT does not fully understand the implicit time in the question.
On the contrary, GenTKGQA performs well in both simple and complex question types due to the fact that we use a deep manner to incorporate subgraph information into the LLM.
However, similar to other LLMs, GenTKGQA randomly generates some irrelevant answers when generating multiple answers, e.g., in the "Simple Entity" question type.


\section{Conclusion}

We propose a novel generative framework, GenTKGQA, which guides the LLM in a two-stage manner to handle temporal question answering on TKG.
Specifically, at the subgraph retrieval phase, we exploit the LLM's intrinsic knowledge to mine the temporal constraints and structural links in the temporal questions, which reduces the search space of the subgraphs in both temporal and structural dimensions. 
We employ the in-context learning approach to complete subgraph retrieval for the entire dataset with a small number of samples.
In order to improve the inference performance of the LLM on complex question types, at the answer generation phase, we present the instruction tuning technique to make the open-source LLM truly understand the temporal order and structural dependencies among retrieved facts. 
Most significantly, we design novel virtual knowledge indicators to establish a bridge between subgraph neural information and text representations. 
Experimental results show that our framework can effectively utilize the LLM to solve the complex question type of TKGQA task and validate the adequacy of our motivation.


\section*{Limitations}

Although the complex temporal question on the CronQuestions dataset involves multiple facts, the inter-entity connection in each fact is single-hop, so the hyper-parameter $k$ of our model is set to 1 to achieve the best results. 
In fact, the vast majority of current TKGQA datasets involve facts that are single-hop. 
So, in the future, we will explore more datasets to solve inference for multi-hop complex temporal problems over TKG.
In addition, we use the in-context learning approach to prompt the ChatGPT baseline to answer the questions, saving the labor cost of checking whether the answers are correct. However, the design of different templates may result in incomplete consistency with the manual results, but this does not affect the conclusions of this paper. Because the results provided by other works similarly show the poor performance of ChatGPT's temporal question answering \cite{TEMPReason,ARI}.

\section*{Ethics Statement}

This work presents a novel two-stage framework for the temporal knowledge graph question answering task using large language models.
Our experiments use the publicly available datasets and language models from open sources.
The dataset is developed to be used for the TKG-based temporal QA task. The language models are used to generate answers to temporal questions with entities or timestamps, which does not involve toxic content.
This paper uses the above dataset and models with their initial intention.
We believe that this work is consistent with ACL's ethics policy and presents no potential risk.

\section*{Acknowledgements}
This work is sponsored in part by the National Natural Science Foundation of China under Grant No. 62025208, and the Xiangjiang Laboratory Fund under Grant No. 22XJ01012.


\bibliography{ref}

\newpage

\appendix

\begin{table}[t]
\centering
	\resizebox{.85\linewidth}{!}{
    \begin{tabular}{lccc}  
    \toprule
	\textbf{Category} & \textbf{Train} & \textbf{Dev} & \textbf{Test} \\ 
    \midrule
    Simple Entity & 90,651 & 7,745 & 7,812 \\
    Simple Time & 61,471 & 5,197 & 5,046 \\
	Before/After & 23,869 & 1,982 & 2,151 \\
    First/Last & 118,556 & 11,198 & 11,159 \\
    Time Join & 55,453 & 3,878 & 3,832 \\
    \midrule
    Simple Reasoning & 152,122 & 12,942 & 12,858 \\
    Complex Reasoning & 197,878 & 17,058 & 17,142 \\
    \midrule
    Entity Answer & 225,672 & 19,362 & 19,524 \\
    Time Answer & 124,328 & 10,638 & 10,476 \\
    \midrule
    \textbf{Total} & 350,000 & 30,000 & 30,000 \\
    \bottomrule
    \end{tabular}}
    \caption{Dataset Statistics of CronQuestions.}
\label{tab:data1}
\end{table}

\begin{table}[t]
\centering
	\resizebox{.65\linewidth}{!}{
    \begin{tabular}{lccc}  
    \toprule
	\textbf{Category} & \textbf{Train} & \textbf{Dev} & \textbf{Test} \\ 
    \midrule
    Explicit & 2,725 & 1,302 & 1,312\\
    Implicit & 660 & 296 & 297 \\
    Temporal & 2,810 & 1,177 & 1,163\\
    Ordinal & 976 & 587 & 593 \\
    \midrule
    \textbf{Total} & 7,171  & 3,362  & 3,365  \\
    \bottomrule
    \end{tabular}}
     \caption{Dataset Statistics of TimeQuestions\protect\footnotemark.}
\label{tab:data2}
\end{table}
\footnotetext{The actual number of questions in the training, validation and test set is 6,970, 3,236 and 3,237, respectively. The total number exceeds the number of questions as some questions belong to multiple categories.}


\section{Dataset Statistics and Metrics}
\label{appendix:dataset}

We use the CronQuestions and TimeQuestions dataset in our experiments. TimeQuestions was first proposed by \citet{TimeQuestions}, but it provides only static knowledge graphs with temporal attributes, not strictly temporal knowledge graphs. Later, \citet{TwiRGCN} expanded this dataset by preprocessing all the contained facts into the temporal knowledge graph format of (subject entity, relationship, object entity, [start time, end time]) and restricting all times to years. Dataset statistics are described in Table \ref{tab:data1} and \ref{tab:data2}, respectively.

Following previous studies, we leverage two popular evaluation metrics, Hits@1 and Hits@10. Specifically, $\text { Hits@K }=\frac{1}{\mid \text {Test}\mid} \sum_{q \in \text { Test }}\operatorname{ind}(\operatorname{rank}(q) \leq \text{K})$, where $\operatorname{rank}(q)$ denotes the ranking of the answer to question $q$ obtained by the model in the candidate list. $\operatorname{ind}$ is 1 if the inequality holds and is 0 otherwise, $\text{K}=1,10$. 

\section{Baselines and Implementation Details}
\label{appendix:baseline}

We use the OpenAI-API\footnote{\url{https://platform.openai.com/docs/api-reference}} (gpt-3.5-turbo-0613\footnote{\url{https://platform.openai.com/docs/models/gpt-3-5-turbo}}) for all ChatGPT-related experiments, including subsequent ChatGPT baselines.

In the subgraph retrieval phase, we use ChatGPT to mine temporal constraints and structural links between entities and add 5 samples to the in-context learning prompt templates, which are presented as Table \ref{tab:rel prompt} and \ref{tab:time prompt}. 
We set $k$=1 for the top-$k$ relations. 
In the answer generation phase, following \cite{TNTComplEx}, we select the dimension of entity/relation/time embeddings to 512.  For T-GNN, the layer $l$ is set to 1, the linear transformations $\mathbf{W}_{q}$, $\mathbf{W}_{k}$ and $\mathbf{W}_{m}$ are 512×512, and the $\mathbf{m}$ and $\mathbf{b}$ of the MLP layer are 512×$|\mathcal{E}|$. 
We use the open-source Llama 2-7B \cite{Llama2} for instruction tuning and select up to $n$=10 relevant facts as additional knowledge. The linear projection layer $\mathbf{W}_{p}$ is 512×4096.
We fine-tune Llama 2-7B using LoRA \cite{lora} with rank 64. The number of epochs is set to 4 and the learning rate is 3e{-4}. We use the AdamW optimizer \cite{Adamw} with a fixed batch size of 8. We conduct all the experiments with NVIDIA A100 GPUs, and the results of each experiment are averaged over three runs.
We will release the source code upon acceptance.

We compare our model with the following baselines:

\textbf{EmbedKGQA} \cite{EmbeddedKGQA}:
Timestamps are ignored during pre-training and random time embeddings are used during the QA task.

\textbf{EaE} \cite{EaE}: 
In the experiment, we follow use TKG embeddings to enhance the question representation, and then predict the answer probabilities via dot-product.

\textbf{CronKGQA} \cite{cronqa}: 
CronKGQA is the TKGQA embedding-based method that first uses a LM model to get question embeddings and then utilize a TKG embedding-based scoring function for answer prediction.

\textbf{EntityQR and TempoQR} \cite{tempoqa}: 
Based on EaE, EntityQR utilizes a TKG embedding-based scoring function for answer prediction.
TempoQR utilizes a TKG embedding-based scoring function for answer prediction and fuse additional temporal information.

\textbf{TMA} \cite{TMA}:
TMA improves QA performance through enhanced fact retrieval and adaptive fusion network.

\textbf{TSQA} \cite{TSQA}:
TSQA presents a contrastive learning module that improves sensitivity to time relation words.

\textbf{TwiRGCN} \cite{TwiRGCN}: 
TwiRGCN is a method for processing TKGQA tasks with the relational graph convolutional network (RGCN).

\textbf{BERT and RoBERTa} \cite{BERT,roberta}: 
For \textbf{\textit{w/o tkg}}, following CronKGQA \cite{cronqa}, we add a prediction head on top of the [CLS] token of the final layer, and then do a softmax over it to predict the answer probabilities.
For \textbf{\textit{w/ tkg}}, following TempoQR \cite{tempoqa}, we generate their LM-based question embedding and concatenate it with the annotated entity and time embeddings, followed by a learnable projection. The resulted embedding is scored against all entities and timestamps via dot-product.

\textbf{ChatGPT}:
To ensure that the output format meets the expected requirements, we use the in-context learning approach to motivate ChatGPT to answer the questions and provide 5 examples in the prompt template. 
The specific templates are presented in Table \ref{tab:chatgpt prompt} of Appendix \ref{appendix:prompt}. 
\textbf{\textit{w/o tkg}} and \textbf{\textit{w/ tkg}} differ in whether or not question-relevant facts are provided in the input prompts, which are retrieved in the first stage by our proposed GenTKGQA framework.

\begin{table*}[ht]
\centering
	\resizebox{.9\textwidth}{!}{
    \begin{tabular}{|l|}  
    \toprule
    \textbf{Relation Ranking Prompt}\\
    \midrule
    l will give you a list of words. \\
    Find the $\{k\}$ words from the list that are most semantically related to the given sentence. \\
    If there are no semantically related words, pick out any $\{k\}$ words. \\
    \\
    Examples)\\
    \\
    Sentence A: When was the first time Martin Taylor played for The Hatters?\\
    Words List: [`member of sports team', `position held', `award received', `spouse', `employer']\\
    Top \{$k$\} Answers: [`member of sports team']\\
    \\
    \dots\\
    \\
    Sentence E: Which was awarded to Daniel Walther in 1980?\\
    Words List: [`member of sports team', `position held', `award received', `spouse', `employer']\\
    Top \{$k$\} Answers: [`award received']\\
    \\
    Now let’s find the top \{$k$\} words. \\
    Sentence: \{$sentence$\} \\
    Words List: \{$relation\_list$\} \\
    Top \{$k$\} Answer:\\
    \bottomrule
    \end{tabular}}
    \caption{Relation Ranking Prompt. This prompt is used to extract structural links between entities in the question.}
\label{tab:rel prompt}
\end{table*}

\begin{table*}[ht]
\centering
    \resizebox{.9\textwidth}{!}{
    \begin{tabular}{|l|}  
    \toprule
    \textbf{Time Mining Prompt}\\
    \midrule
    I will give you a natural language question with a temporal constraint. \\
    Answer the temporal constraint involved in the question based on the knowledge context and the question type.\\
    Answer only in "before", "after", "between and" format.\\
    \\
    Examples)\\
    \\
    Question A: Who held Governor of Connecticut position after Lowell P. Weicker?\\
    Knowledge Context: [`Lowell P. Weicker', `position held', `Governor of Connecticut', `1991', `1995']\\
    Question Type: after\\
    Response: after 1995\\
    \\
    \dots\\
    \\
    Question E: Who's the player who played in AC Reggiana with Daniele Magliocchetti?\\
    Knowledge Context: [`Daniele Magliocchetti', `member of sports team',  `A.C. Reggiana', `2012', `2014']\\
    Question Type: time\_join\\
    Response: between 2012 and 2014\\
    \\
    Next, let's answer the time constraints involved in the following question. \\
    Question: \{$question$\} \\
    Knowledge Context: \{$context$\} \\
    Question Type: \{$type$\}\\
    Response:\\
    \bottomrule
    \end{tabular}}
    \caption{Time Mining Prompt. This prompt is used to find the time constraints involved in the complex question.}
\label{tab:time prompt}
\end{table*}

\begin{table*}[ht]
\centering
    \resizebox{.9\textwidth}{!}{
    \begin{tabular}{|l|}  
    \toprule
    \textbf{Instruction Tuning Template}\\
    \midrule
    Below is an instruction that describes a task, paired with an input that provides further context. \\
    Write a response that appropriately completes the request.\\\\
    Instruction:\\
    Answer the questions based on evidence.\\
    Each evidence is in the form of [head, relation, tail, start\_time, end\_time]\\ and it means `head relation is tail between start\_time and end\_time'.\\
    You must list the 10 most relevant answers.\\\\
    Input:\\
    Question: \{$question$\} \\
    Evidence set: \{$evidence\_set$\}\\\\
    Response:\{$answer$\}\\
    \bottomrule
    \end{tabular}}
    \caption{This is the template for instruction tuning.}
\label{tab:instruction tuning}
\end{table*}

\begin{table*}[ht]
\centering
    \resizebox{.9\textwidth}{!}{
    \begin{tabular}{|l|}  
    \toprule
    \textbf{ChatGPT \textit{w/ tkg}}\\
    \midrule
    Answer the questions based on evidence.\\
    Each evidence is in the form of [head, relation, tail, start\_time, end\_time]\\ and it means `head relation is tail between start\_time and end\_time'.\\
    You must list the 10 most relevant answers separated by `$\textbackslash$t'.\\
    \\
    Examples)\\
    \\
    Question A: Who was the Member of the House of Representatives in 1990?\\
    Evidence set: [[`Simon Crean', `position held', `Member of the House of Representatives', `1990', `2013'],\dots]\\
    Answer: Simon Crean$\textbackslash$tJohn Dawkins$\textbackslash$t\dots\\
    \\
    \dots\\
    \\
    Question E: With whom did Steve Haslam play on the Sheffield Wednesday F.C.?\\
     Evidence set: [[`Ola Tidman', `member of sports team', `Sheffield Wednesday F.C.', `2003', `2005'], \dots]\\
    Answer: Ola Tidman$\textbackslash$tChris Marsden$\textbackslash$t\dots\\
    \\
    Now let's answer the Question based on the Evidence set. \\
    Please do not say there is no evdience, you must list the 10 most relevant answers separated by `$\textbackslash$t'.\\
    Question: \{$question$\} \\
    Evidence set: \{$evidence\_set$\}\\
    Answer:\\
    \midrule
    \textbf{ChatGPT \textit{w/o tkg}}\\
    Answer the questions directly.\\
    You must answer the 10 most relevant answers separated by `$\textbackslash$t'.\\
     \\
    Examples)\\
    \\
    Question A: Who was the Member of the House of Representatives in 1990?\\
    Answer: Simon Crean$\textbackslash$tJohn Dawkins$\textbackslash$t\dots\\
    \\
    \dots\\
    \\
    Question E: With whom did Steve Haslam play on the Sheffield Wednesday F.C.?\\
    Answer: Ola Tidman$\textbackslash$tChris Marsden$\textbackslash$t\dots\\
    \\
    Now let's answer the Question, you must answer the 10 most relevant answers separated by `$\textbackslash$t'.\\
    Question: \{$question$\} \\
    Answer:\\
    \bottomrule
    \end{tabular}}
    \caption{ChatGPT Baseline Prompt.}
\label{tab:chatgpt prompt}
\end{table*}

\section{Prompt Template} \label{appendix:prompt}

The prompts for relation ranking and time mining can be found in Table \ref{tab:rel prompt} and Table \ref{tab:time prompt}, respectively.
The template used for instruction tuning is shown in Table \ref{tab:instruction tuning}. 
The ChatGPT baseline prompt is presented in Table \ref{tab:chatgpt prompt}.

\end{document}